\documentclass[lettersize,journal]{IEEEtran}
\usepackage{amsmath,amsfonts}
\usepackage{algorithmic}
\usepackage{array}
\usepackage[caption=false,font=normalsize,labelfont=sf,textfont=sf]{subfig}
\usepackage{textcomp}
\usepackage{stfloats}
\usepackage{url}
\usepackage{verbatim}
\usepackage{graphicx}
\usepackage{multirow}
\usepackage{makecell} % For \multirowcell command
\usepackage{hyperref}
\usepackage[justification=centering]{caption}
\usepackage[dvipsnames]{xcolor}

\def\BibTeX{{\rm B\kern-.05em{\sc i\kern-.025em b}\kern-.08em
    T\kern-.1667em\lower.7ex\hbox{E}\kern-.125emX}}
\usepackage{balance}
\begin{document}
\title{A Dual-Domain Convolutional Network for Hyperspectral Single-Image Super-Resolution}
\author{Murat Karayaka, Usman Muhammad, Jorma Laaksonen, Md Ziaul Hoque and Tapio Seppänen
\thanks{Manuscript created Dec, 2025; This research work is financially supported by the Center for Machine Vision and Signal Analysis, University of Oulu, Finland.
\\
\indent M. Karayaka, M.Z. Hoque and T. Seppänen are with Center for Machine Vision and Signal Analysis, University of Oulu. Usman Muhammad and Jorma Laaksonen are with the Department of Computer Science, Aalto University, Espoo, Finland.}}

\markboth{Under peer review}%
{How to Use the IEEEtran \LaTeX \ Templates}

\maketitle

\begin{abstract}
This study presents a lightweight dual-domain super-resolution network (DDSRNet) that combines Spatial-Net with the discrete wavelet transform (DWT). Specifically, our proposed model comprises three main components: (1) a shallow feature extraction module, termed Spatial-Net, which performs residual learning and bilinear interpolation; (2) a low-frequency enhancement branch based on the DWT that refines coarse image structures; and (3) a shared high-frequency refinement branch that simultaneously enhances the LH (horizontal), HL (vertical), and HH (diagonal) wavelet subbands using a single CNN with shared weights.  As a result, the DWT enables subband decomposition, while the inverse DWT reconstructs the final high-resolution output. By doing so, the integration of spatial- and frequency-domain learning enables DDSRNet to achieve highly competitive performance with low computational cost on three hyperspectral image datasets, demonstrating its effectiveness for hyperspectral image super-resolution. The source codes are publicly available at: \href{https://github.com/mkarayak24/DDSRNet} {https://github.com/mkarayak24/DDSRNet} 

\end{abstract}


\section{Introduction}

\IEEEPARstart{H}{}yperspectral images (HSIs) provide rich spectral information but are limited in spatial resolution due to hardware constraints, while multispectral images (MSIs) achieve higher spatial resolution with fewer spectral bands \cite{li2019hyperspectral}. Single image super-resolution (SR) aims to reconstruct a high-resolution (HR) image from a low-resolution (LR) observation. Earlier SR methods predominantly used interpolation techniques such as nearest-neighbor, bilinear, and bicubic interpolation \cite{jiang2020single}. Although computationally efficient and well suited to the hardware of their time, these approaches operated purely on pixel-level computations without incorporating contextual or prior information, resulting in HR images that often lacked fine structural details \cite{li2019hyperspectral}.

In recent years, deep learning–based methods have become dominant in the remote sensing domain \cite{muhammad2018pre, muhammad2019bag, muhammad2018feature,muhammad2022patch}, especially in super-resolution tasks, \cite{chudasama2024comparison, muhammad2025dacn}, demonstrating remarkable improvements in both reconstruction accuracy and perceptual quality. In particular, single-image super-resolution methods have become more popular compared to fusion-based approaches which require an additional RGB or panchromatic image \cite{10406185}. The main reason is that single-image super-resolution methods are more practical for real-time deployment. Thus, single-image super-resolution methods have received increased attention in the literature. For instance, a lightweight depthwise separable dilated convolutional neural network that integrates mean squared error (MSE) along with spectral angle–based loss functions was proposed in \cite{11073307}. Meanwhile, a spectral–spatial fusion approach that employs an Inception-inspired architecture followed by a dedicated multi-scale fusion block has been introduced in \cite{11074908}. However, the depthwise separable dilated network may suffer from limited global context and spectral distortion, while the Inception-like fusion model introduces higher computational complexity and risks of overfitting on small datasets.

To enhance both spatial resolution and spectral integrity, a plug-and-play module called spectral–spatial unmixing fusion (SSUF) was proposed in \cite{11187306}. The channel-attention-based spatial–spectral feature extraction network was found to be highly effective in preserving spectral fidelity, as reported in \cite{zhang2024hyperspectral}. A spatial–spectral interactive Transformer (SSIT) block integrated into both the encoder and decoder of a U-Net architecture was employed in \cite{xu20233}. Furthermore, a novel Group-Autoencoder (GAE) framework combined with a diffusion model to construct high-quality images was introduced in \cite{wang2024enhancing}. However, these methods often suffer from high computational complexity and require extensive training data to achieve stable performance.

We are motivated by the observation that low-frequency components (e.g., smooth structures and global intensity variations) and high-frequency components (e.g., edges and fine textures) exhibit fundamentally different characteristics \cite{lu2017wavelet}. Conventional CNNs often process these components together in the spatial domain, which can result in blurred edges or over-smoothing. By transforming the upsampled features into the wavelet domain, we can explicitly separate these components, thereby improving reconstruction quality while reducing computational cost. Therefore, we propose a lightweight dual-domain super-resolution network (DDSRNet) for hyperspectral single image super resolution that combines Spatial-Net with the discrete wavelet transform (DWT). 

Specifically, the purpose of the Spatial-Net is to extract and enhance spatial features from the low-resolution input, providing a strong spatial representation for subsequent wavelet-domain refinement. It learns structural and contextual cues such as edges, shapes, and textures, which are crucial for preserving spatial fidelity in the reconstructed high-resolution image. On the other hand, the purpose of the Discrete Wavelet Transform (DWT) is to decompose the feature maps into low- and high-frequency components, allowing the network to process smooth structures and fine textures separately. This separation helps preserve edge details, reduce over-smoothing, and improve the overall reconstruction quality of the hyperspectral image. Finally, a hybrid loss function is designed that combines the reconstruction loss in the image domain with auxiliary losses computed in the wavelet and the spatial domains. 

In summary, the contributions of this work are summarized as:

\begin{enumerate}
    \item We present a novel dual-domain model, called DDSRNet, that operates in both the spatial and wavelet domains, with each component designed to be lightweight and suitable for real-time deployment.
    \item We employ a hybrid loss function that jointly optimizes spatial reconstruction and frequency-domain consistency.
    \item Experiments conducted on three hyperspectral datasets under $2\times$, $4\times$, and $8\times$ downsampling scenarios demonstrate that the proposed method achieves highly competitive performance across all cases.
\end{enumerate}

%\section{Related Work}

\section{Methodology}
Fig. 1 illustrates the overall architecture of the proposed DDSRNet model, which comprises two key components: (1) a Spatial-Net that performs initial feature learning, and (2) a Wavelet Decomposition (DWT) module that refines the reconstruction through frequency-domain processing. We first describe the Spatial-Net operating in the spatial domain, followed by the Discrete Wavelet Transform (DWT) module in the frequency domain. The subsequent sections provide a detailed analysis of each component, along with the proposed hybrid loss function used to optimize the overall model.

\begin{figure*}[htbp]
    \centering
    \includegraphics[width=0.86\textwidth]{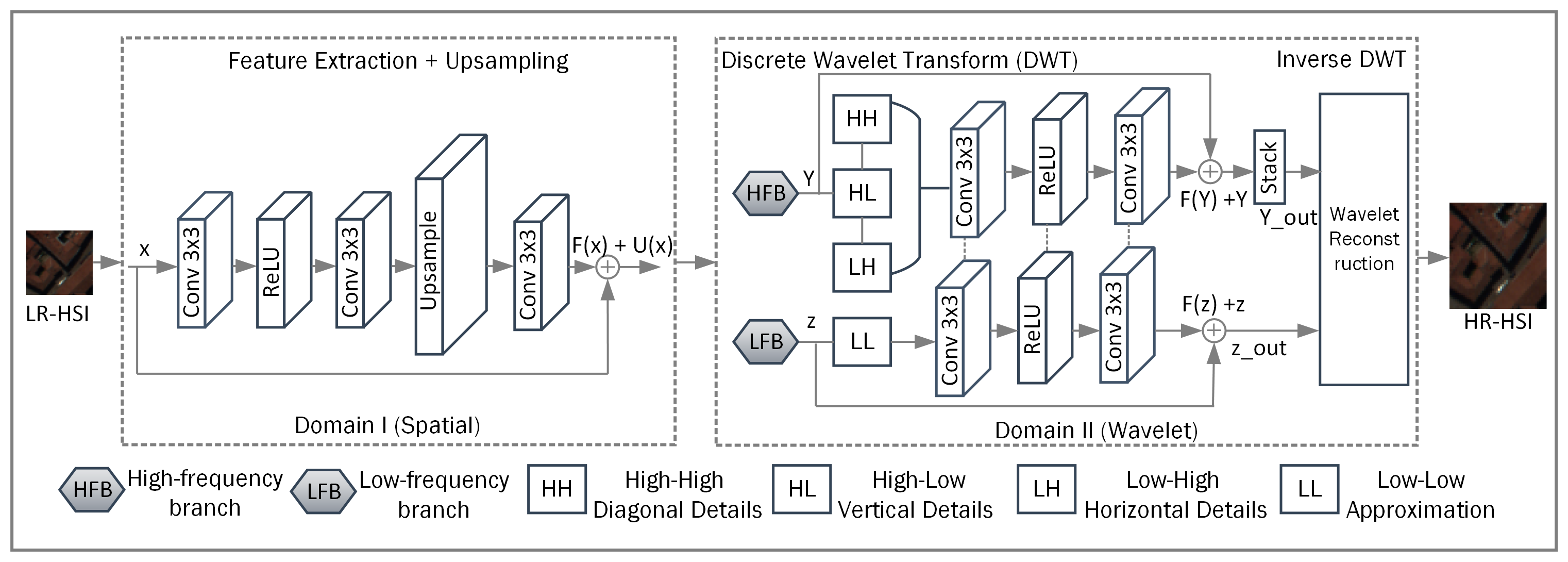}
    \caption{An overview of the proposed DDSRNet model. The left block represents the first domain, i.e., the Spatial-Net, while the right block illustrates the procedure of the Discrete Wavelet Transform (DWT).}
    \label{fig:model}
\end{figure*}

\subsection{Spatial-Net}
The Spatial-Net  consists of two successive convolutional layers with ReLU activation, a non-learnable bilinear upsampling layer, and a final projection convolution. 

Mathematically, given an input tensor $X \in \mathbb{R}^{B\times C\times H\times W}$, where $B$ is the batch size, $C$ the number of channels, and $H\times W$ the spatial size, Spatial-Net computes:
\begin{equation}
F = \mathrm{Conv}_2\!\big(\sigma(\mathrm{Conv}_1(X))\big),
\end{equation}
where $F \in \mathbb{R}^{B\times C_h\times H\times W}$ is the low-resolution feature map after two convolutions. $\sigma(\cdot)$ denotes the ReLU function. Similarly, we then apply bilinear upsampling followed by a $3{\times}3$ convolution:

\begin{equation}
Y_{\text{main}} = \mathrm{Conv}_3\!\big(U(F)\big),
\end{equation}
where $U(\cdot)$ is bilinear upsampling by factor $s$, and $Y_{\text{main}} \in \mathbb{R}^{B\times C\times sH\times sW}$ is the learned high-resolution path output, and $C_h$ is the hidden width:

\begin{equation}
\hat{Y} = Y_{\text{main}} + U(X), \qquad
\hat{Y} \in \mathbb{R}^{B\times C\times sH\times sW}.
\label{eq:spatialnet-out}
\end{equation}
where $\hat{Y}$ denotes the Spatial-Net output, and
$U(X)\in\mathbb{R}^{B\times C\times sH\times sW}$ is the bilinear skip connection.

\subsection{Discrete Wavelet Decomposition (DWT)}
\label{subsec:dwt}
After the image is processed by the Spatial-Net, we apply a discrete wavelet transform (DWT) to decompose the upsampled feature map into low and high frequency components \cite{lu2017wavelet}. This process explicitly separates global structural information from fine textures, allowing the network to learn more effective refinements for both components. In our implementation, the DWT is performed using the Haar wavelet, which applies separable low-pass and high-pass filtering along both spatial dimensions, followed by downsampling.

Mathematically, let $\mathbf{X}_{\uparrow} \in \mathbb{R}^{B \times C \times sH \times sW}$ denote the upsampled input, where $B$ is the batch size, $C$ is the number of channels, and $(sH, sW)$ are the spatial dimensions after upsampling. The 2-D DWT produces one low-frequency subband $\mathbf{Y}_{\mathrm{LL}}$ and three high-frequency subbands $\mathbf{Y}_{\mathrm{LH}}$, $\mathbf{Y}_{\mathrm{HL}}$, and $\mathbf{Y}_{\mathrm{HH}}$, obtained as \cite{lu2017wavelet}:
\begin{equation}
\mathbf{Y}_{\mathrm{LL}},\ \mathbf{Y}_{\mathrm{LH}},\ \mathbf{Y}_{\mathrm{HL}},\ \mathbf{Y}_{\mathrm{HH}}
= \mathrm{DWT}(\mathbf{X}_{\uparrow})
\end{equation}
where $\mathbf{Y}_{\mathrm{LL}}$ represents the coarse, low-pass features, while $\mathbf{Y}_{\mathrm{LH}}$, $\mathbf{Y}_{\mathrm{HL}}$, and $\mathbf{Y}_{\mathrm{HH}}$ capture vertical, horizontal, and diagonal high-frequency components, respectively. The resulting tensors have dimensions $[B, C, H_d, W_d]$ for $\mathbf{Y}_{\mathrm{LL}}$
and $[B, C, 3, H_d, W_d]$ for the stacked high-frequency components, where
$H_d = \tfrac{s H_0}{2}$ and $W_d = \tfrac{s W_0}{2}$,
with $H_0$ and $W_0$ denoting the original input height and width, respectively.

In the model, the low-frequency subband $\mathbf{Y}_{\mathrm{LL}}$ is processed by a low frequency branch (LFB), which refines smooth regions and overall image structure using residual convolutional layers. Meanwhile, the high-frequency branch (HFB), represented as $\mathbf{Y}_{\mathrm{H}} = [\mathbf{Y}_{\mathrm{LH}}, \mathbf{Y}_{\mathrm{HL}}, \mathbf{Y}_{\mathrm{HH}}]$, is enhanced by a shared Wavelet Branch, where each subband is refined through a small convolutional block with a residual connection to restore textures and sharpen edges. This design allows the model to learn orientation-specific features in the wavelet domain efficiently.

Finally, the reconstructed image is obtained via the inverse wavelet transform (IDWT):
\begin{equation}
\mathbf{X}_{\mathrm{SR}} = \mathrm{IDWT}\!\left(\mathbf{Y}_{\mathrm{LL}}^{\ast},\, \mathbf{Y}_{\mathrm{H}}^{\ast}\right).
\end{equation}
where $\mathbf{Y}_{\mathrm{LL}}^{\ast}$ and $\mathbf{Y}_{\mathrm{H}}^{\ast}$ are the refined low-frequency and high-frequency feature maps. This explicit frequency separation and recombination improve edge fidelity and structural consistency while minimizing oversmoothing in the reconstructed high-resolution output.

\subsection{Hybrid Loss Function}
\label{subsec:loss}

A hybrid loss function is employed to jointly guide spatial- and frequency-domain learning by combining image-domain reconstruction loss with auxiliary spatial and wavelet losses. Specifically, the hybrid loss is based on the Huber loss, which combines the strengths of mean squared error (MSE) and mean absolute error (MAE). The first component of the hybrid loss function is the main reconstruction loss,
defined as the Huber loss between the high-resolution ground truth
$\mathbf{Y}_{\text{HR}}$ and the model prediction $\mathbf{Y}_{\text{pred}}$:
\begin{equation}
\mathcal{L}_{\text{rec}} =
\text{Huber}(\mathbf{Y}_{\text{pred}}, \mathbf{Y}_{\text{HR}}).
\end{equation}

The second component is the spatial loss, defined as the Huber loss between the
high-resolution ground truth $\mathbf{Y}_{\text{HR}}$ and the Spatial-Net output
$\mathbf{Y}_{\text{pred,spatial}}$:
\begin{equation}
\mathcal{L}_{\text{spatial}} =
\text{Huber}(\mathbf{Y}_{\text{pred,spatial}}, \mathbf{Y}_{\text{HR}}).
\end{equation}

In addition, the ground-truth image is decomposed into low- and high-frequency
subbands using the discrete wavelet transform (DWT).
The corresponding model outputs, $\mathbf{Y}^{\text{pred}}_{\mathrm{L}}$ and
$\mathbf{Y}^{\text{pred}}_{\mathrm{H}}$, are compared with their ground-truth
counterparts, $\mathbf{Y}_{\mathrm{L}}$ and
$\mathbf{Y}_{\mathrm{H}} = [\mathbf{Y}_{\mathrm{LH}},
\mathbf{Y}_{\mathrm{HL}}, \mathbf{Y}_{\mathrm{HH}}]$, to enforce accurate
reconstruction across both frequency domains:
\begin{equation}
\mathcal{L}_{\text{low}} =
\text{Huber}(\mathbf{Y}^{\text{pred}}_{\mathrm{L}}, \mathbf{Y}_{\mathrm{L}}),
\qquad
\mathcal{L}_{\text{high}} =
\text{Huber}(\mathbf{Y}^{\text{pred}}_{\mathrm{H}}, \mathbf{Y}_{\mathrm{H}}).
\end{equation}

The total training objective combines all four components as
\begin{align}
\mathcal{L}_{\text{total}} =
&\ \lambda_{\text{rec}} \mathcal{L}_{\text{rec}}
+ \lambda_{\text{spatial}} \mathcal{L}_{\text{spatial}} \nonumber \\
&+ \lambda_{\text{low}} \mathcal{L}_{\text{low}}
+ \lambda_{\text{high}} \mathcal{L}_{\text{high}},
\end{align}
where the $\lambda$ coefficients control the relative importance of the individual loss terms. This hybrid formulation encourages the network to produce spatially accurate reconstructions while enhancing edge sharpness and texture fidelity through frequency-domain supervision.

\begin{table}[htbp]
\caption{Ablation results of quantitative performance on the PaviaU dataset at scale 4 with model complexity.}
\label{tab:ablation_study}
\centering
\resizebox{0.75\columnwidth}{!}{
\begin{tabular}{|l|c|c|}
%\hline
\multicolumn{3}{c}{\textbf{Ablation Study}} \\
\hline
\textbf{Model Variant} & \textbf{MPSNR$\uparrow$} & \textbf{SAM$\downarrow$} \\
\hline
Without Spatial-Net                    & 30.218 & 4.932 \\
Without Wavelet-Net                    & 30.398 & 4.829 \\
Without Shared Wavelet Branch          & 30.534 & 4.853 \\
Without Band Grouping                  & 30.524 & \textbf{4.731} \\
Without Hybrid Loss                    & 30.541 & 4.845 \\
%Overlapping Patches            & 30.369 & 4.906 \\
\textbf{DDSRNet (Full Model)}     & \textbf{30.561}  & 4.836 \\
\hline
%\hline
\multicolumn{3}{c}{\textbf{Model Complexity}} \\
\hline
\textbf{Model} & \textbf{Scale} & \textbf{Parameters} \\ 
\hline
ERCSR \cite{li2021exploring}             & 4x & 1.59M \\
MCNet \cite{lii2020mixed}                & 4x & 2.17M \\
PDENet \cite{hou2022deep}                & 4x & 2.30M \\
CSSFENet \cite{zhang2024hyperspectral}   & 4x & 1.61M \\
\textbf{DDSRNet} (Ours)           & 4x & 0.07M \\
\hline
\end{tabular}
}
\end{table}

\begin{figure}[h]
    \centering
    \includegraphics[width=0.80\columnwidth]{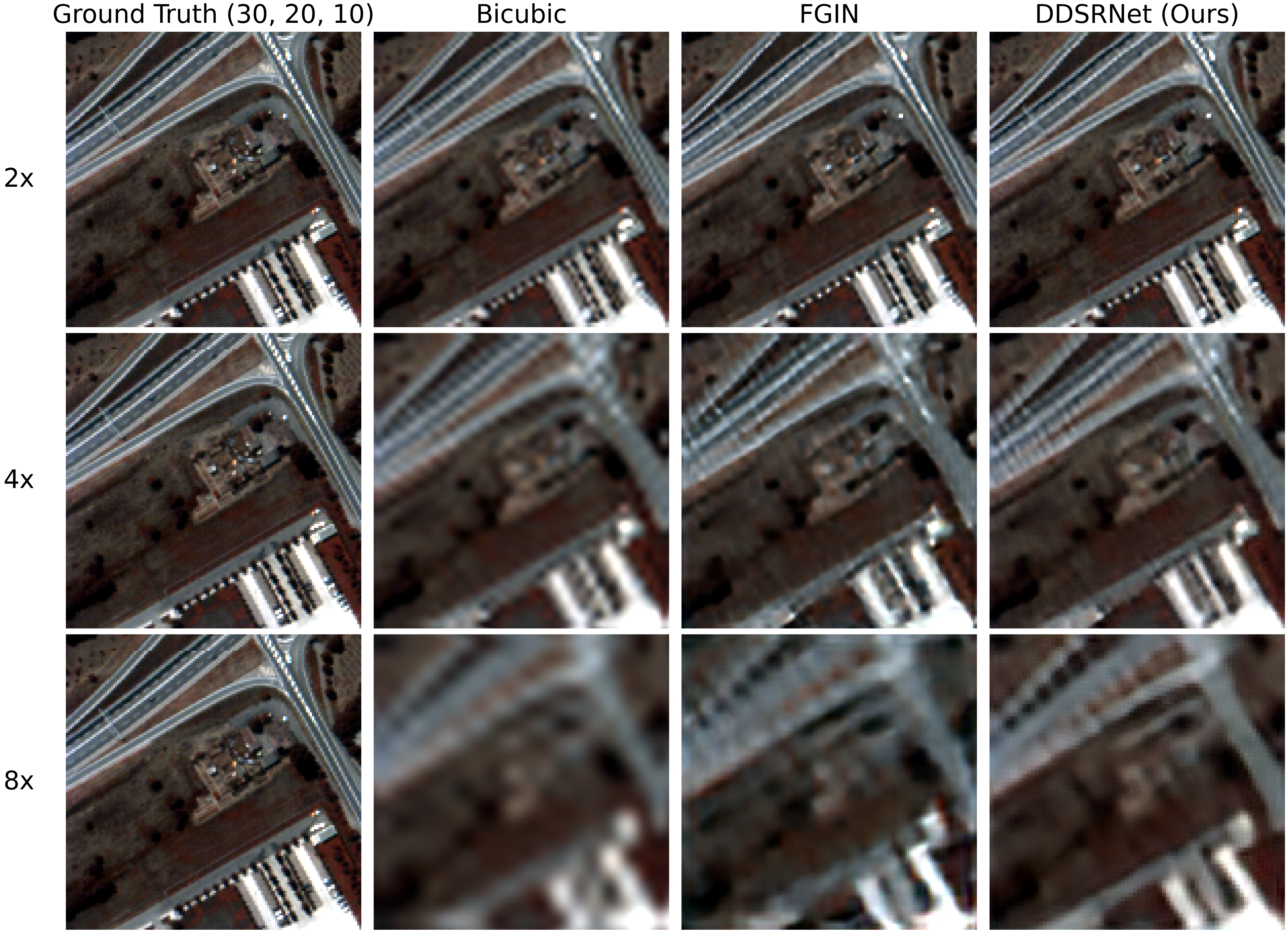}
    \caption{Qualitative comparison on the PaviaU test image (false-color composite) at scaling factors of 2×, 4×, and 8×.}
    \label{fig:comparison}
\end{figure}

\begin{table*}
\centering
\caption{Evaluation on datasets (PaviaC, PaviaU) under different scaling factors. Results are reported from \cite{zhang2024hyperspectral}.}
\label{tab:quantitative_results}
\resizebox{0.6\textwidth}{!}{%
\renewcommand{\arraystretch}{0.9} % Adjust row height
\small
\begin{tabular}{|c|c|ccc|ccc|}
\hline
\multirow{2}{*}{\textbf{Scale Factor}} & \multirow{2}{*}{\textbf{Model}} 
& \multicolumn{3}{c|}{\textbf{PaviaC}} 
& \multicolumn{3}{c|}{\textbf{PaviaU}} \\ 
\cline{3-8}
& & \textbf{MPSNR$\uparrow$} & \textbf{MSSIM$\uparrow$} & \textbf{SAM$\downarrow$} 
  & \textbf{MPSNR$\uparrow$} & \textbf{MSSIM$\uparrow$} & \textbf{SAM$\downarrow$} \\ \hline

% ---------- 2x ----------
\multirow{9}{*}{$\boldsymbol{2\times}$} 
& VDSR \cite{kim2016accurate}              & 34.87 & 0.9501 & 3.689 & 34.03 & 0.9524 & 3.258 \\ 
& EDSR \cite{lim2017enhanced}              & 34.58 & 0.9452 & 3.898 & 33.98 & 0.9511 & 3.334 \\ 
& MCNet \cite{lii2020mixed}                & 34.62 & 0.9455 & 3.865 & 33.74 & 0.9502 & 3.359 \\ 
& MSDformer \cite{chen2023msdformer}       & 35.02 & 0.9493 & 3.691 & 34.15 & 0.9553 & 3.211 \\ 
& MSFMNet \cite{zhang2021multi}            & 35.20 & 0.9506 & 3.656 & 34.98 & 0.9582 & 3.160 \\ 
& AS3 ITransUNet \cite{xu20233}            & 35.22 & 0.9511 & 3.612 & 35.16 & 0.9591 & 3.149 \\ 
& PDENet \cite{hou2022deep}                & 35.24 & 0.9519 & 3.595 & 35.27 & 0.9594 & 3.142 \\ 
& CSSFENet \cite{zhang2024hyperspectral}   & 35.52 & 0.9544 & 3.542 & 35.92 & \textbf{0.9625} & 3.038 \\  
& \textbf{DDSRNet}                  & \textbf{36.39} & \textbf{0.9601} & \textbf{3.288} &  \textbf{36.43} & 0.9550 & \textbf{2.921} \\ 
\hline

% ---------- 4x ----------
\multirow{9}{*}{$\boldsymbol{4\times}$} 
& VDSR \cite{kim2016accurate}              & 28.31 & 0.7707 & 6.514 & 29.90 & 0.7753 & 4.997 \\ 
& EDSR \cite{lim2017enhanced}              & 28.59 & 0.7782 & 6.573 & 29.89 & 0.7791 & 5.074 \\ 
& MCNet \cite{lii2020mixed}                & 28.75 & 0.7826 & 6.385 & 29.99 & 0.7835 & 4.917 \\ 
& MSDformer \cite{chen2023msdformer}       & 28.81 & 0.7833 & 5.897 & 30.09 & 0.7905 & 4.885 \\ 
& MSFMNet \cite{zhang2021multi}            & 28.87 & 0.7863 & 6.300 & 30.28 & 0.7948 & 4.861 \\ 
& AS3 ITransUNet \cite{xu20233}            & 28.87 & 0.7893 & 5.972 & 30.28 & 0.7940 & 4.859 \\ 
& PDENet \cite{hou2022deep}                & 28.95 & 0.7900 & 5.876 & 30.29 & 0.7944 & 4.853 \\ 
& CSSFENet \cite{zhang2024hyperspectral}   & 29.05 & 0.7961 & 5.816 & \textbf{30.68} & 0.8107 & 4.839 \\ 
& \textbf{DDSRNet}                  & \textbf{29.56} & \textbf{0.8232} & \textbf{5.542} & 30.56 & \textbf{0.8181} & \textbf{4.836} \\ 
\hline

% ---------- 8x ----------
\multirow{9}{*}{$\boldsymbol{8\times}$} 
& VDSR \cite{kim2016accurate}              & 24.80 & 0.4944 & 7.588 & 27.02 & 0.5962 & 7.133 \\ 
& EDSR \cite{lim2017enhanced}              & 25.06 & 0.5282 & 7.507 & 27.46 & 0.6302 & 6.678 \\ 
& MCNet \cite{lii2020mixed}                & 25.09 & 0.5391 & 7.429 & 27.48 & 0.6254 & 6.683 \\ 
& MSDformer \cite{chen2023msdformer}       & 25.21 & 0.5462 & 7.427 & 27.32 & 0.6341 & 6.668 \\ 
& MSFMNet \cite{zhang2021multi}            & 25.25 & 0.5464 & 7.449 & 27.58 & 0.6356 & 6.615 \\ 
& AS3 ITransUNet \cite{xu20233}            & 25.25 & 0.5435 & 7.417 & 27.68 & 0.6413 & 6.574 \\ 
& PDENet \cite{hou2022deep}                & 25.28 & 0.5436 & 7.402 & 27.73 & 0.6457 & 6.531 \\ 
& CSSFENet \cite{zhang2024hyperspectral}   & 25.35 & 0.5493 & \textbf{7.306} & \textbf{27.82} & \textbf{0.6569} & \textbf{6.505} \\ 
& \textbf{DDSRNet}                  & \textbf{25.57} & \textbf{0.5847} & 7.627 & 27.64 & 0.6250 & 6.770 \\\hline

\end{tabular}%
}
\end{table*}

\begin{table}[tb!]
\centering
\caption{Evaluation on the Chikusei dataset in different scaling setups. The comparison results are sourced from \cite{wang2024enhancing}.}
\resizebox{0.35\textwidth}{!}{
\begin{tabular}{|c|c|c|c|c|}
\hline
\multirow{1}{*}{\centering \textbf{Scale}} &
\multirow{1}{*}{\centering \textbf{Model}} &
\textbf{MPSNR$\uparrow$} &
\textbf{MSSIM$\uparrow$} &
\textbf{SAM$\downarrow$} \\ \hline
\multirow{8}{*}{\centering $\boldsymbol{2\times}$} 
    & Bicubic     & 35.008 & 0.932 & 1.718 \\ 
    & EDSR    \cite{lim2017enhanced}    & 35.489 & 0.941 & 2.444 \\ 
    & GDRRN   \cite{li2018single}    & 35.958 & 0.939 & 1.561 \\ 
    & SSPSR   \cite{jiang2020learning}    & 35.723 & 0.944 & 2.275 \\ 
    & MCNet    \cite{lii2020mixed}   & 36.371 & 0.948 & 1.784 \\ 
    & GELIN    \cite{wang2022group}   & 37.747 & 0.959 & 1.384 \\ 
    & DIFF     \cite{wang2024enhancing}   & \textbf{38.748}  & \textbf{0.966}  & 1.638 \\ 
    & \textbf{DDSRNet}        & 38.406 & 0.963 & \textbf{1.044}  \\
    \hline
\multirow{8}{*}{\centering $\boldsymbol{4\times}$} 
    & Bicubic     & 29.676 & 0.770 & 3.161 \\ 
    & EDSR    \cite{lim2017enhanced}    & 29.976 & 0.799 & 4.127 \\ 
    & GDRRN   \cite{li2018single}     & 30.658 & 0.801 & 2.913 \\ 
    & SSPSR    \cite{jiang2020learning}   & 30.858 & 0.823 & 3.196 \\ 
    & MCNet     \cite{lii2020mixed}   & 31.189 & 0.821 & 2.955 \\ 
    & GELIN     \cite{wang2022group}   & 31.095 & 0.838 & 2.834 \\ 
    & DIFF       \cite{wang2024enhancing} & 32.248 & \textbf{0.860}  & 3.507 \\ 
    & \textbf{DDSRNet}      & \textbf{32.528}  & 0.859 &  \textbf{2.146}  \\ 
    \hline
\end{tabular}
}
\label{tab:results_chikusei}
\end{table}

\section{Experimental setup}

\subsection{Datasets}
The performance of the model was evaluated on three benchmark hyperspectral datasets: Pavia Center (PaviaC) with 102 spectral bands, Pavia University (PaviaU) with 103 bands, and Chikusei, which provides a higher spectral resolution of 128 bands. % These datasets were selected to ensure a comprehensive evaluation across distinct acquisition conditions, spatial characteristics, and spectral dimensionalities.

\subsection{Implementation and Evaluation Metrics}
To utilize the PaviaU and PaviaC datasets, training and testing samples for PaviaU and PaviaC were generated by extracting 
$144 \times 144$ patches, following \cite{zhang2024hyperspectral}. In the Chikusei dataset, we first cropped a $512 \times 512$ region from the center of the image to exclude black areas. Subsequently, we extracted $64 \times 64$ patches for the $2\times$ downscaling setting and $128 \times 128$ patches for the $4\times$ downscaling setting, as suggested in \cite{wang2024enhancing}. Finally, all patches were downsampled with scale factors of $2\times$, $4\times$, and $8\times$, and the model was trained to reconstruct the original high-resolution images from these degraded inputs.

In addition, we extracted overlapping patches from the PaviaU dataset using a stride of $18\times18$, while keeping non-overlapping patches for the other datasets. Spectral band grouping was applied with a group size of $35$ without overlaps. We padded the training data to $105$ channels by duplicating the last band as needed (e.g., three additional bands for PaviaC and two for PaviaU) to ensure equal channel sizes across datasets. For the Chikusei dataset, which originally contained $128$ bands, we padded the data to $140$ channels. During the evaluation, the padded channels were removed and only the original spectral bands were compared. For the PaviaU and PaviaC datasets, we adopted the same test patch as in \cite{zhang2024hyperspectral}; in the case of the Chikusei dataset, the top left patch was used as the test set. The remaining patches were used for training and validation. The models were trained using Adam Optimizer with a learning rate of $0.0001$ and a batch size of $4$. Training was performed for up to $6000$ epochs, with early stopping based on validation loss and patience of $200$ epochs. A single-level Haar wavelet decomposition was applied to decouple spatial frequency components, and we allocated equal weights of $0.35$ to each term in the hybrid loss function. As part of our evaluation of the test set, we applied widely used quantitative metrics, including Mean Peak Signal-to-Noise Ratio (MPSNR), Mean Structural Similarity Index (MSSIM), Spectral Angle Mapper (SAM), Root Mean Squared Error (RMSE), and Cross-Correlation (CC).

\subsection{Ablation Study}

We conducted an ablation study on the PaviaU dataset with a $4\times$ downscaling factor to evaluate the individual contribution of each component in the proposed model. Table~\ref{tab:ablation_study} summarizes the results in terms of two HSI quality metrics, including MPSNR and SAM. We quantitatively evaluate the contribution of each component in both domains, such as Spatial-Net and Wavelet-Net. The replacement of Spatial-Net with a standard bilinear upsampling layer led to a significant drop in MPSNR by $0.343$ dB and an increase in SAM by $0.096^\circ$. Removing the Wavelet-Net and directly using the Spatial-Net output as the final prediction also resulted in a decrease in MPSNR by $0.163$ dB, although SAM decreased slightly by $0.007^\circ$. This highlights the advantage of frequency-domain processing. We further examined the effect of replacing the shared wavelet branch with independent branches for each high-pass sub-band, which negatively impacted all metrics.

Additionally, removing band grouping reduced MPSNR by $0.037$ dB and increased model complexity, although it improved SAM by $0.105^\circ$. Finally, introducing the Hybrid Loss function led to a further improvement in MPSNR by $0.02$ dB and a reduction in SAM error by $0.09^\circ$. When all components are integrated, the full DDSRNet model achieves the best overall performance, demonstrating the complementary effect of each module.

We further demonstrate the efficiency of the proposed model in terms of complexity. Specifically, DDSRNet requires only $0.07$M parameters, which is significantly fewer than CSSFENet \cite{zhang2024hyperspectral}. Figure \ref{fig:comparison} presents a qualitative comparison on the PaviaU test image among bicubic interpolation, FGIN \cite{11074908}, and the proposed DDSRNet. Under the challenging $8\times$ scaling scenario, bicubic interpolation produces over-smoothed and blurred results, while FGIN preserves sharper structures but introduces mild aliasing artifacts. In contrast, DDSRNet effectively reconstructs fine edges and textures with fewer distortions. Overall, the proposed framework achieves an excellent trade-off between reconstruction accuracy and efficiency, making it well-suited for deployment in resource-constrained environments.

\subsection{Comparison with State-of-the-Art Methods}

The proposed DDSRNet model is evaluated on the PaviaC and PaviaU datasets under $2\times$, $4\times$, and $8\times$ downscaling factors, with quantitative results summarized in Table~\ref{tab:quantitative_results}, where DDSRNet shows superior performance under the $2\times$ setting.
For the more challenging $4\times$ scaling factor, our model outperformed CSSFENet~\cite{zhang2024hyperspectral} on PaviaC, gaining $0.51$ dB in MPSNR, $0.0271$ in MSSIM, and reducing SAM by $0.274$. On PaviaU, it showed similar performance to CSSFENet, with almost identical SAM and slightly higher MSSIM, while showing only a small drop of $0.12$ dB in MPSNR. At the $8\times$ downscaling factor, DDSRNet had the highest MPSNR on PaviaC, surpassing CSSFENet by $0.22$ dB, and showed better performance in MSSIM, while having a slightly higher SAM value than CSSFENet. On PaviaU, its performance was close to CSSFENet, with only minor differences ($-0.18$ dB in MPSNR, $-0.0319$ in MSSIM, and $+0.265$ in SAM).

Table~\ref{tab:results_chikusei} presents the experimental results across the $2\times$ and $4\times$ downscaling factors on the Chikusei dataset. At the $2\times$ scaling factor, DDSRNet achieved an MPSNR of $38.406$ dB, which is $0.342$ dB lower than the diffusion-based DIFF model~\cite{wang2024enhancing}, but higher than all other CNN-based models. DDSRNet also obtained an MSSIM of $0.963$, nearly identical to the DIFF model’s $0.966$. In addition, it yielded a SAM value of $1.044$. For the $4\times$ downscaling factor, DDSRNet achieved the highest MPSNR of $32.528$ dB, outperforming the DIFF model by $0.28$ dB, while producing a comparable MSSIM ($0.859$ vs.\ $0.860$). Moreover, the SAM was reduced to $2.146$, representing a decrease of $1.36$ compared to the best reference method. These results confirm that DDSRNet outperforms recent approaches in addressing complex HSI super-resolution tasks.

\section{Conclusion}

In this study, we introduced a dual-domain convolutional neural network for hyperspectral image super-resolution. The proposed architecture combines a Spatial-Net for initial feature extraction with a Discrete Wavelet Transform (DWT) module for frequency-domain refinement. In addition, a hybrid Huber loss-based loss function further enhances the reconstruction accuracy by jointly optimizing image, spatial and frequency domain components. Experimental results on three benchmark hyperspectral datasets demonstrate that our model achieves competitive reconstruction quality. In future work, we plan to extend the framework toward transformer-based or diffusion-driven dual-domain architectures to further improve generalization and efficiency across diverse hyperspectral imaging scenarios.

\bibliographystyle{IEEEtran}
\bibliography{references.bib}

\end{document}